\documentclass[a4paper, 10pt, conference]{ieeeconf}      % Use this line for a4 paper

\IEEEoverridecommandlockouts                              % This command is only needed if 
\usepackage{graphics} % for pdf, bitmapped graphics files
\usepackage[pdftex]{graphicx}
\usepackage{amsmath} % assumes amsmath package installed
\usepackage{amssymb}  % assumes amsmath package installed

\newtheorem{define}{Definition}
\title{\LARGE \bf
Circular FREAK-ORB Visual Odometry
}

\author{Daniel J. Mankowitz$^{1}$ Ehud Rivlin$^{2}$% <-this % stops a space
\thanks{$^{1}$Daniel J. Mankowitz is with the Department of Electrical Engineering,
Technion Israel Institute of Technology
        {\tt\small danielm@tx.technion.ac.il}}%
\thanks{$^{2}$Ehud Rivlin is with the Department of Computer Science,
Technion Israel Institute of Technology
        {\tt\small ehudr@cs.technion.ac.il}}%
}

\begin{document}

\maketitle
\thispagestyle{empty}
\pagestyle{empty}

%%%%%%%%%%%%%%%%%%%%%%%%%%%%%%%%%%%%%%%%%%%%%%%%%%%%%%%%%%%%%%%%%%%%%%%%%%%%%%%%
\begin{abstract}
We present a novel Visual Odometry algorithm entitled Circular FREAK-ORB (CFORB). This algorithm detects features using the well-known ORB algorithm \cite{c31} and computes feature descriptors using the FREAK algorithm \cite{c8}. CFORB is invariant to both rotation and scale changes, and is suitable for use in environments with uneven terrain. Two visual geometric constraints have been utilized in order to remove invalid feature descriptor matches. These constraints have not previously been utilized in a Visual Odometry algorithm. A variation to circular matching \cite{c12} has also been implemented. This allows features to be matched between images without having to be dependent upon the epipolar constraint. This algorithm has been run on the KITTI benchmark dataset and achieves a competitive average translational error of $3.73 \%$ and average rotational error of $0.0107 deg/m$. CFORB has also been run in an indoor environment and achieved an average translational error of $3.70 \%$. After running CFORB in a highly textured environment with an approximately uniform feature spread across the images, the algorithm achieves an average translational error of $2.4 \%$ and an average rotational error of $0.009 deg/m$.
\end{abstract}

%%%%%%%%%%%%%%%%%%%%%%%%%%%%%%%%%%%%%%%%%%%%%%%%%%%%%%%%%%%%%%%%%%%%%%%%%%%%%%%%
\section{Introduction}
\label{sec:introduction}

%1. The Visual Odometry Problem
Visual Odometry (VO) involves estimating the ego-motion of an object using the images from single or multiple cameras as an input. VO is a fully passive approach and offers a cheaper and mechanically easier-to-manufacture solution compared to Lidar and Radar sensors. 
%2. Why is this important
The ability to estimate the ego-motion of an object is crucial in many applications where accurate localization sensors such as GPS are not available. One such application is indoor navigation where an agent is required to navigate to a specified location, without the use of GPS, in order to perform a task \cite{b00, b0}. This is crucial for navigating in environments such as warehouses, hospitals and museums to name a few.

Another application involves ego-motion estimation for planetary landing in space exploration tasks\cite{b1}. VO has also been utilised on the Mar's rovers to track its motion,  perform slip checks, monitor vehicle safely whilst driving near obstacles as well as approaching various targets and objects efficiently \cite{b2,b3}. This includes performing complex approaches towards targets and obstacles. VO has been used on Unmanned Aerial Vehicles to perform tasks which include point-to-point navigation as well as autonomous take-off and landing \cite{b4}. Another important application of VO is in underwater robots \cite{b5}. These robots do not have access to or cannot rely on GPS coordinates. The ocean floor is generally rich in texture and therefore provides a good environment for robust feature extraction which is a key requirement to performing VO.

VO is an active field and numerous works have been published on this topic. Both monocular and stereo algorithms have been developed. Real-time Topometric Localization performs monocular VO using features called Whole Image SURF (WI-SURF) descriptors \cite{c9}. This method is prone to occlusion errors but exhibits good performance over large distances. Another monocular algorithm detects features using the Kanade-Lucas-Tomasi feature detector and requires only a single correspondence to perform ego-motion estimation which is computationally efficient \cite{c17}. Among the stereo algorithms,  Multi-frame Feature Integration detects features using a combination of Harris Detectors and FREAK descriptors. The features are tracked using an optical flow algorithm. Ego-motion is estimated by minimizing the re-projection errors \cite{b000}. This algorithm exhibits very good VO performance on the well-known KITTI dataset. StereoScan is another method that detects features using custom-made corner and blob masks. Features are matched using their Sobel responses and a further circular matching procedure is performed in order to determine the features to be used for ego-motion estimation \cite{c12}.

%There are a number of well-known datasets, most notably the KITTI dataset \cite{c4}, that have been used as performance benchmarks. 

%3. Possible/current solution(s) to the VO problem

%4. Our solution
Both \cite{b000} and \cite{c12} have inspired the development of our algorithm entitled Circular FREAK-ORB. This algorithm detects features using an ORB detector and matches features using  FREAK descriptors. A variation of circular matching, similar to that using in \cite{c12}, has been utilized to select the correct features for ego-motion estimation and novel geometrical constraints, that have not previously been utilized in VO applications, are utilized to remove incorrect matches.

Therefore the main technical contributions in this work are three-fold:
\begin{itemize}
\item The development of a novel algorithm CFORB that builds on the basis of two previously successful VO algorithms \cite{b000, c12}. This algorithm is invariant to both rotation and translation.
\item Visual geometrical constraints that have not previously been utilized in a VO application to remove bad matches
\item A variation to the circular matching procedure performed by StereoScan \cite{c12} which removes the requirement of utilizing the epipolar constraint to match features between images.
\end{itemize} 
%Technical contributions

\section{Background}

%Visual Odometry 

%ORB Feature Detector
\subsection{ORB Detector}
\label{sec:orb}
Oriented FAST and Rotated BRIEF (ORB) is a feature extraction algorithm that detects features using the FAST detection method with an added orientation component \cite{c31}. The FAST method typically detects corner features using an intensity threshold which compares the intensity of the central pixel to a circular ring of pixels around the central pixel. FAST does not contain an orientation component. ORB adds an orientation component to FAST by computing the \textit{Intensity Centroid (IC)} of the corner feature that has been detected. The IC assumes that the intensity of a corner is offset from the corner itself and this can be used to determine a patch orientation \cite{c31}. The IC is calculated according to ($\ref{eqn:ic}$). 

\begin{equation}
m_{pq} = \sum_{x,y}x^py^qI(x,y) \enspace ,
\label{eqn:ic}
\end{equation} 

where $m_{pq}$ is the $p,q^{th}$ moment of a patch. This is then utilized to calculate the intensity centroid as 

\begin{equation}
C= \biggl(\frac{m_{10}}{m_{00}}, \frac{m_{01}}{m_{00}}\biggr) \enspace .
\label{eqn:centroid}
\end{equation}

By constructing a vector from the corner to the IC, it is possible to determine the orientation $\theta$ of the patch as 

\begin{equation}
\theta = atan2(m_{10}, m_{01}) \enspace . 
\label{eqn:orientation}
\end{equation}

The moments are computed within a radius $r$ from the corner feature. This radius defines the patch size. This improves rotation invariance \cite{c31}. Once the features have been detected, descriptors need to be computed and these descriptors then need to be matched. 
%FREAK Descriptors

\subsection{FREAK Descriptors}
\label{sec:freak}
The Fast Retina Keypoints (FREAK) descriptor is a relatively new and efficient feature descriptor \cite{c8}. FREAK shares similarities with BRISK descriptors \cite{c3} as it also makes use of a sampling pattern around the detected feature in order to generate the feature descriptor. However, this sampling pattern is inspired by the human retina and contains a dense sampling pattern in the immediate vicinity of the detected feature corresponding to the fovea. This is used to capture high-resolution information. In the outer vicinity of the detected feature, the sampling pattern becomes more sparse corresponding to the perifoveal region. This region is used to detect lower resolution, coarse information. It is these points that help determine whether a pair of descriptors match or not.\\

The first $16$ Bytes of the FREAK descriptor contain the coarse information and if these bytes  for a pair of descriptors are not within a pre-defined threshold of one another, then the match is discarded. This leads to efficient matching which outperforms BRISK, SURF and SIFT by multiple orders of magnitude. It is invariant to both rotation and scale.\\

Matching of the FREAK descriptors can then be performed using a method such as the Hamming distance \cite{c3}.

\subsection{Ego-motion Estimation}
\label{sec:egomotion}
Once the features have been detected and matched between corresponding images, the ego-motion estimation of the cameras can then be computed. One way to compute the ego-motion estimation is by minimizing the sum of re-projection errors as shown in (\ref{eqn:minimize}). 

The feature points detected in the left and right image frames at time $t-1$ are triangulated into 3D and then reprojected into the current image frame at time $t$. The objective function is the sum of reprojection errors and a minimization is performed according to the equation below to obtain the rotation matrix $r$ and the translation vector $t$. 

\begin{equation}
\sum_{i=1}^N \Vert x_i^{(l)}-\pi^{(l)}(\mathbf(X_i);\mathbf{r},\mathbf{t})\Vert^2 + \Vert x_i^{(r)}-\pi^{(r)}(\mathbf(X_i);\mathbf{r},\mathbf{t})\Vert^2 
\label{eqn:minimize}
\end{equation}

Here, $x_i^{(l)}$ and $x_i^{(r)}$ denote the feature locations in the left and right images respectively. $\pi^{(l)}$ and $\pi^{(r)}$ denote the left and right image reprojections respectively. 

\subsection{Epipolar Geometry and Rectification}
\label{sec:epipolar}
\begin{figure*}
\centering
\begin{tabular}{ccc}
\includegraphics[width=0.2\textwidth]{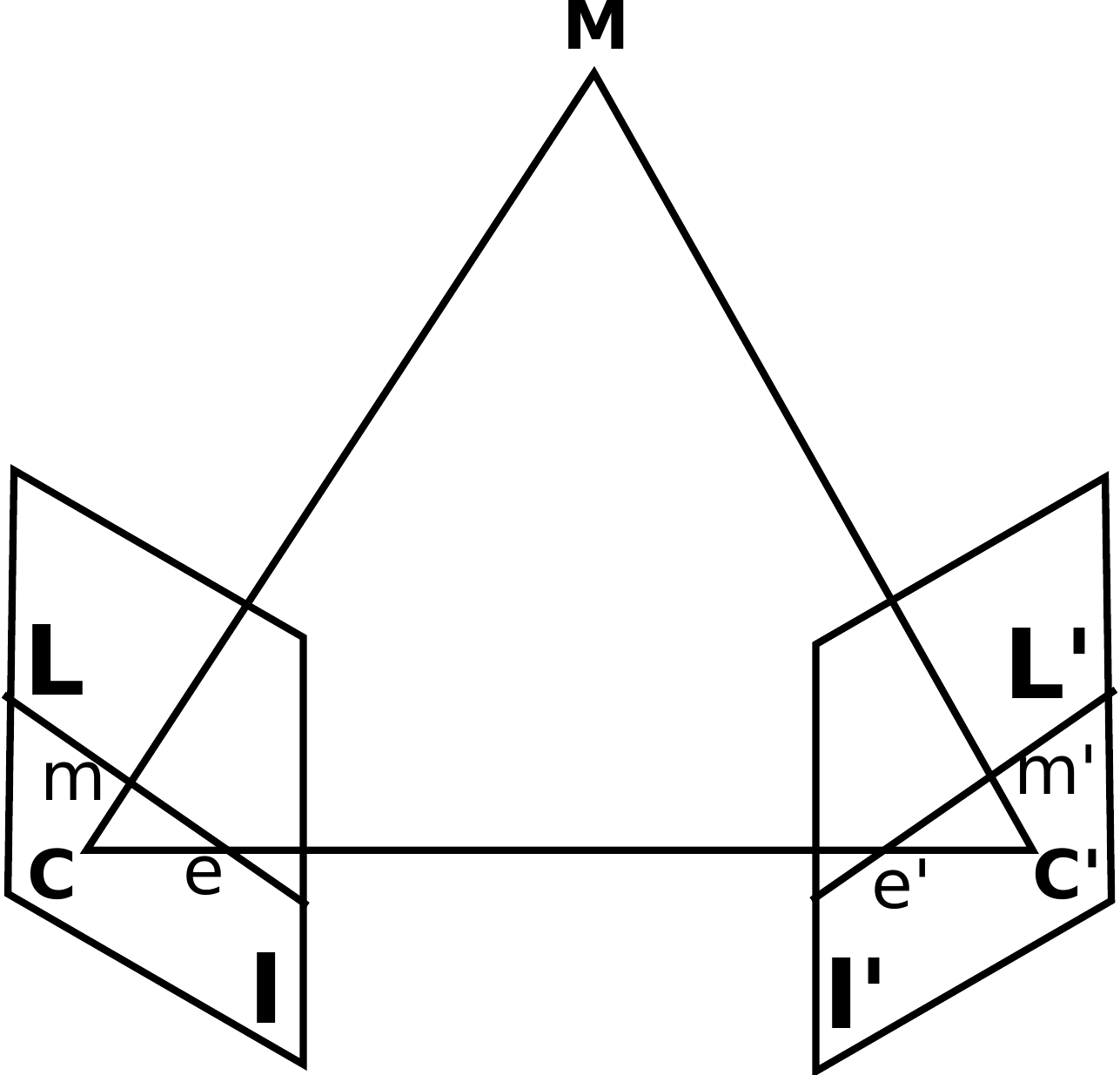} &
\includegraphics[width=0.2\textwidth]{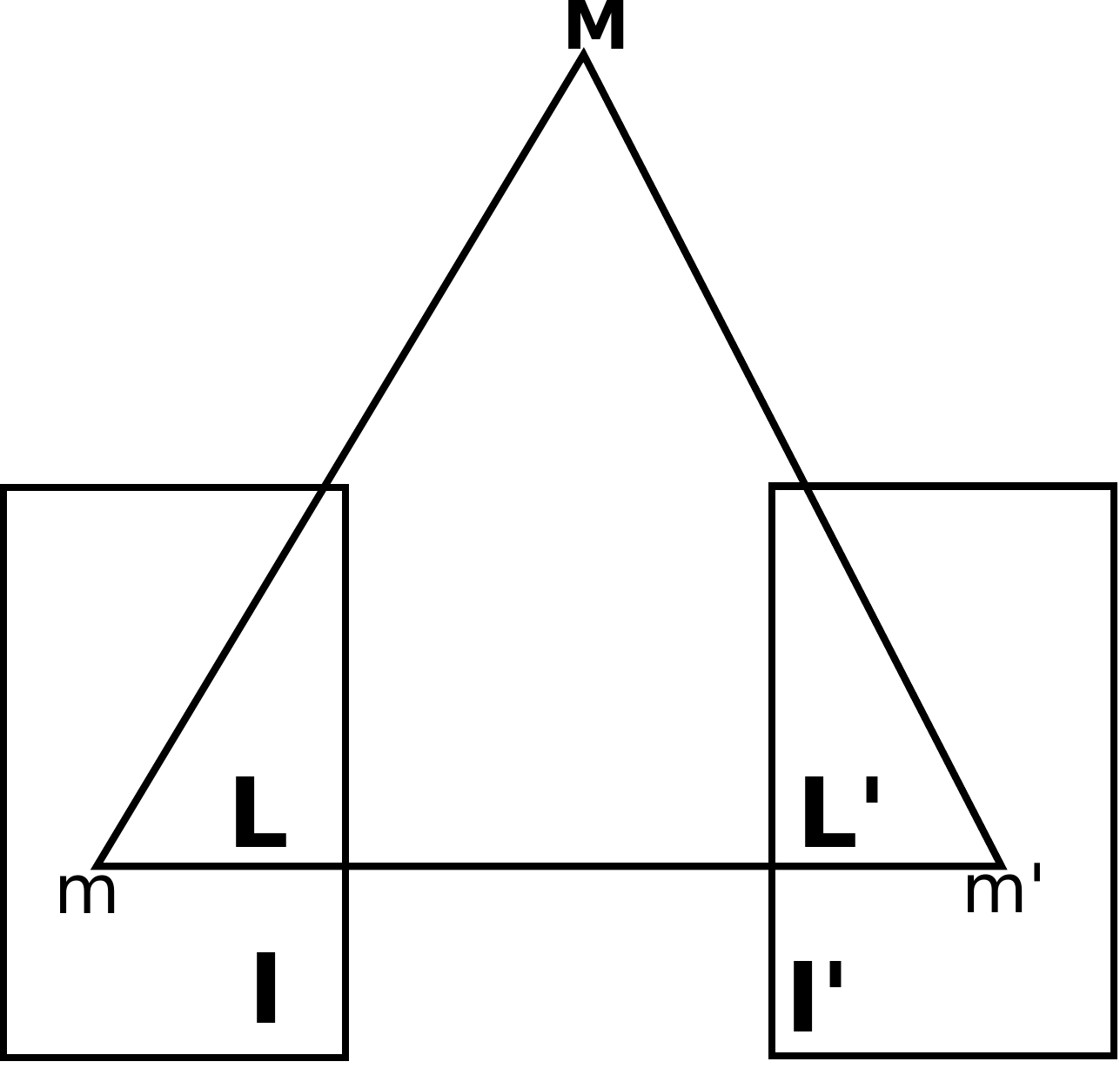} &
\includegraphics[width=0.5\textwidth]{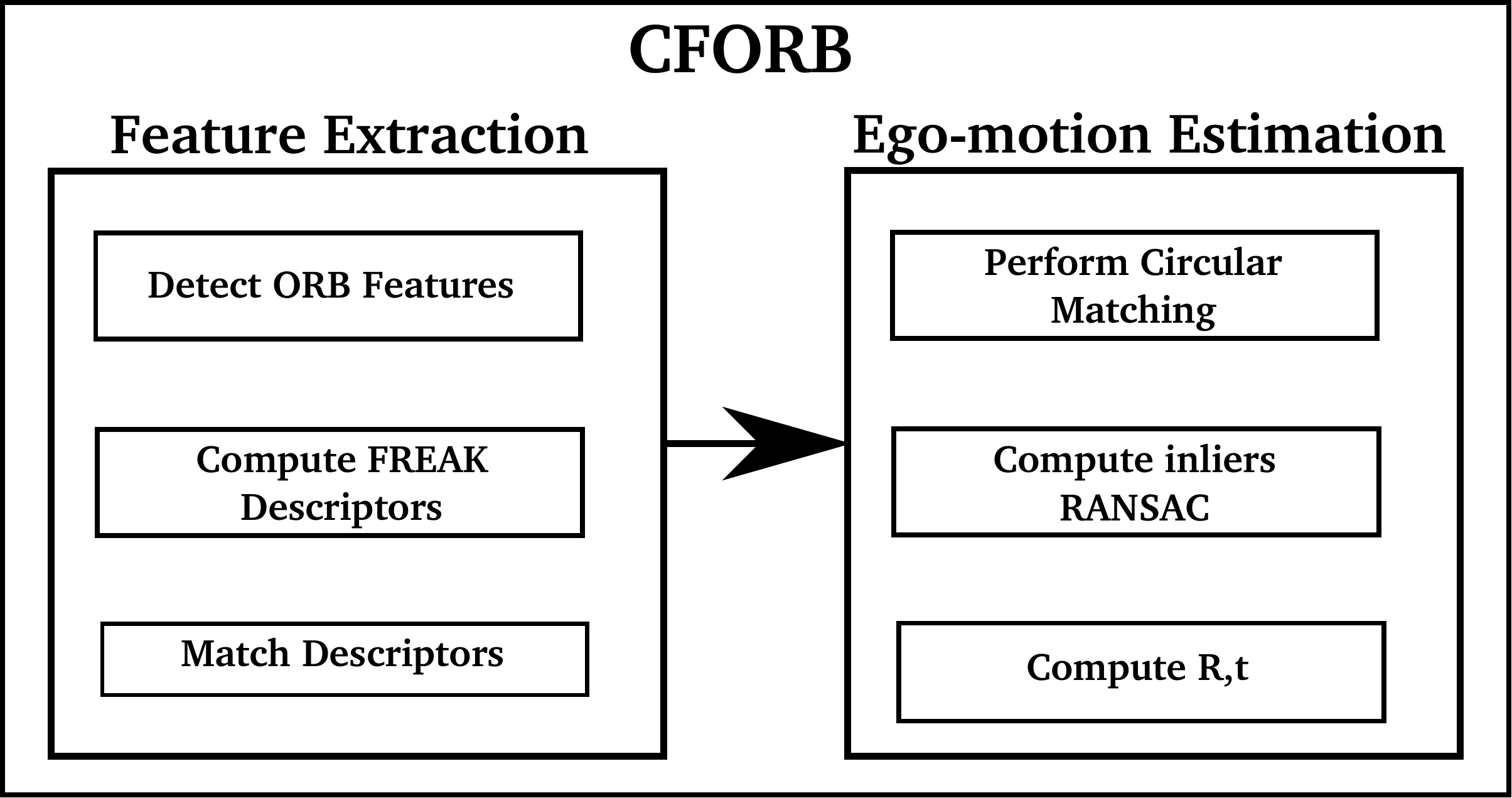}\\
($a$) & ($b$)  &($c$)
\end{tabular}
\caption{Epipolar Geometry: ($a$) Standard setup ($b$)  Rectification setup ($c$) The CFORB Algorithm Overview}
\label{fig:rect}
\end{figure*}

Epipolar geometry provides a complete description of the relative camera geometry between two cameras \cite{b6} as seen in Figure \ref{fig:rect}$a$. The line adjoining the two optical centers, $C$ and $C'$, intersects the images $I$ and $I'$ at their respective epipoles $e$ and $e'$. The projection of the 3D point $M$ onto each image plane is denoted $m$ for image $I$ and $m'$ for image $I'$ respectively. The line $L$ that intersects $m$ and $e$ in image $I$ is called the epipolar line. The corresponding line in image $I'$ is denoted $L'$. Given a point $m$ in image $I$, the corresponding point in image $I'$ is constrained to lie on the epipolar line $L'$. The epipolar constraint is defined as:

\begin{equation}
\textbf{m}'^T \textbf{F} \textbf{m} = 0 \enspace ,
\label{eqn:fund}
\end{equation}

where \textbf{F} is the fundamental matrix.

Image rectification allows for a simplification of the epipolar constraint where the epipolar lines are made to be parallel with the image axis as seen in Figure \ref{fig:rect}$b$ \cite{b7}. A point $m$ in image $I$ is constrained to lie along the epipolar line in the same row in image $I'$. This allows for simplified matching of points between images.

%\begin{figure}[t!]
%\centering
%\includegraphics[width=0.5\textwidth]{../Drawings/cforb.png}
%\caption{CFORB Algorithmic Overview}
%\label{fig:cforb}
%\end{figure}

\section{CFORB Overview}
\label{sec:algorithm}
%Discuss the high level process and any useful additions that have been added.
The CFORB algorithm, presented in Figure \ref{fig:rect}$c$ consists of two major components, namely feature extraction and ego-motion estimation. First, feature extraction is performed on both of the camera images. This includes detecting the features, computing the relevant descriptors and finally matching the descriptors between each of the corresponding stereo images. This also includes visual geometric constraints to weed out bad matches. The ego-motion estimation component performs circular matching similar to StereoScan \cite{c12}, computes the inlier circular features and then estimates the rotation matrix and translation vector which is used to update the location of the camera.

\subsection{Feature Extraction}

\begin{define}
Let $x_l$ and $x_r$ be defined as the \textit{detected} 2D features for the left image $I_t$ and the right image $I'_t$ at time $t$ respectively. Let $\pi_l$ and $\pi_r$ be defined as the corresponding \textit{predicted} 2D features. These features have been (1) triangulated into 3D from the previous image pair at time $t-1$ and (2) projected into the image pair at time $t$. An inlier is defined as a feature that satisfies:

\begin{equation}
(x_l - \pi_l)^2 + (x_r - \pi_r)^2 < \theta \enspace ,
\end{equation}

where $\theta$ is the inlier threshold. A set of corresponding features $\langle x_l, x_r, \pi_l, \pi_r \rangle$ satisfying this constraint are said to be repeatable.
\label{def:inlier}
\end{define}

\subsubsection{Feature Detection}
\label{sec:detection}
There are a number of possible feature detectors that may be utilized for CFORB. The detectors to be compared are BRISK and ORB. We chose these detectors for comparison based on their computational efficiency. The advantage of an ORB detector is the fact that it is invariant to both rotation and scale changes. The rotation invariance comes as a result of computing Intensity Centroids as mentioned in Section \ref{sec:orb}. This is crucial in many VO applications since the camera trajectory may not necessarily be smooth \cite{b2,b5}.\\

\begin{figure}[!t]
\centering
\includegraphics[width=2.5in]{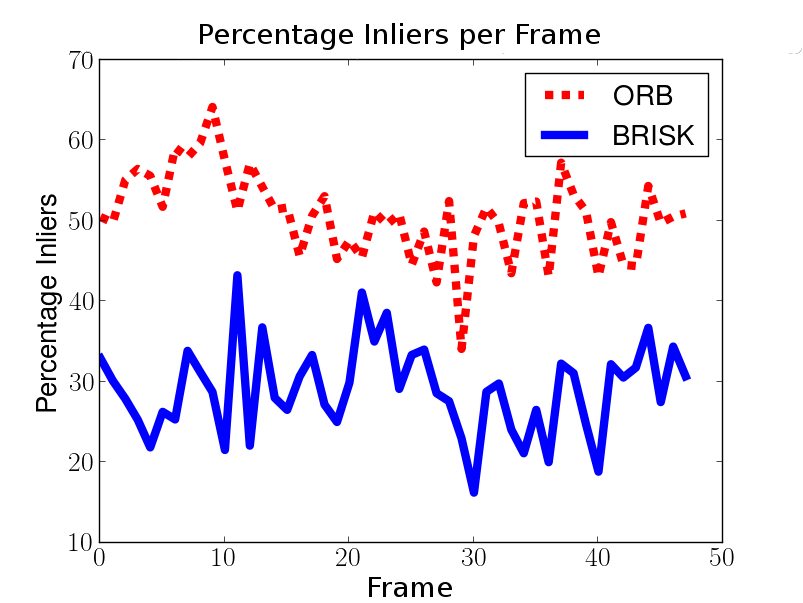}
\caption{The percentage of inliers per image frame on a subset of the KITTI dataset. The inliers are used to estimate the ego-motion of the stereo cameras}
\label{fig:inliers}
\end{figure}

The BRISK detector, however, may exhibit better performance than the ORB detector due to its superior computational efficiency. The BRISK detector is indeed faster than ORB, but is more sensitive to changes in rotation \cite{c31}. In addition, ORB features are more repeatable than BRISK features. To verify this, we ran two variations of the CFORB algorithm on a subset of the KITTI dataset using BRISK and ORB detectors respectively. We compared the inlier percentage for each detector. The inlier percentage indicates the percentage of detected features that are within a pre-defined inlier threshold as defined in Definition \ref{def:inlier}. The percentage of inliers is higher for ORB features compared to BRISK features as seen in Figure \ref{fig:inliers}. An ORB detector therefore results in more repeatable features as well as accurate ego-motion estimates and has therefore been chosen as the feature detection algorithm for CFORB. \\

\subsubsection{Feature Descriptors and Matching}
\label{sec:descriptors}
%Feature descriptos
Descriptors for the ORB features are then computed using the FREAK algorithm. FREAK is multiple orders of magnitude faster than BRISK and ORB descriptors and therefore is an intuitive descriptor choice \cite{c8}. Since FREAK descriptors are generated from multiple binary brightness comparisons, the descriptors can be efficiently matched using the Hamming distance.

\subsubsection{Geometrical Constraints}
\label{sec:constraints}
\begin{figure}[!t]
\centering
\includegraphics[width=3.5in]{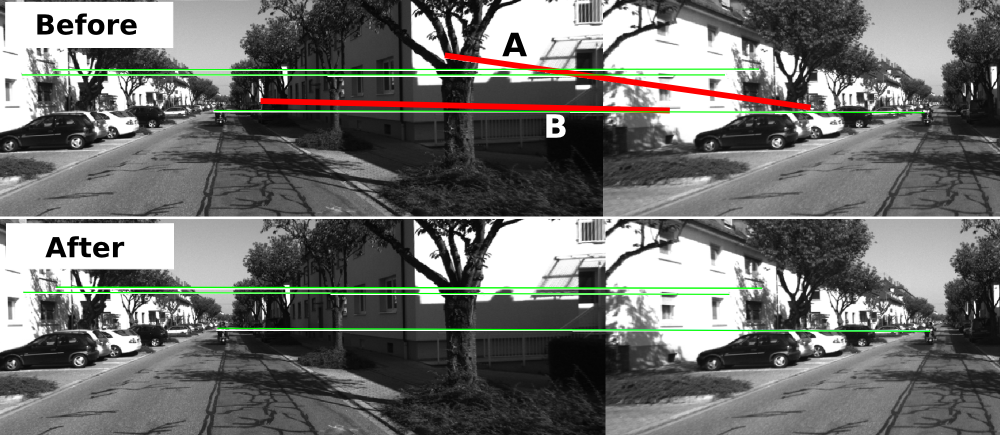}
\caption{The result of applying the vertical and horizontal geometric constraints to remove bad matches}
\label{fig:constraints}
\end{figure}

%weeding out
When matching FREAK descriptors, there were numerous incorrect matches. We have found it possible to remove many of these matches using two visual geometric constraints \cite{c32}. These constraints, namely the vertical and horizontal constraints, have not previously been utilized in a VO algorithm. An example of these constraints in action is presented in Figure \ref{fig:constraints}. The vertical constraint assumes that matches between corresponding images lie on approximately the same row. This assumes that there is not a great deal of rotation between the stereo cameras. The constraint removes a match between corresponding images when the vertical pixel difference between features connected via the red matching line (marked `A' in the figure) is above a pre-defined threshold. The horizontal constraint assumes that matches lie in approximately the same column which is perfectly reasonable in stereo images. Therefore a matching line (marked `B' in the figure) that is significantly below or above the pre-defined pixel difference threshold is removed. Matches that are not removed using these constraints are then passed forward to the ego-motion estimation module.

\subsection{Ego-motion Estimation}
\begin{figure}[!t]
\centering
\includegraphics[width=3.0in]{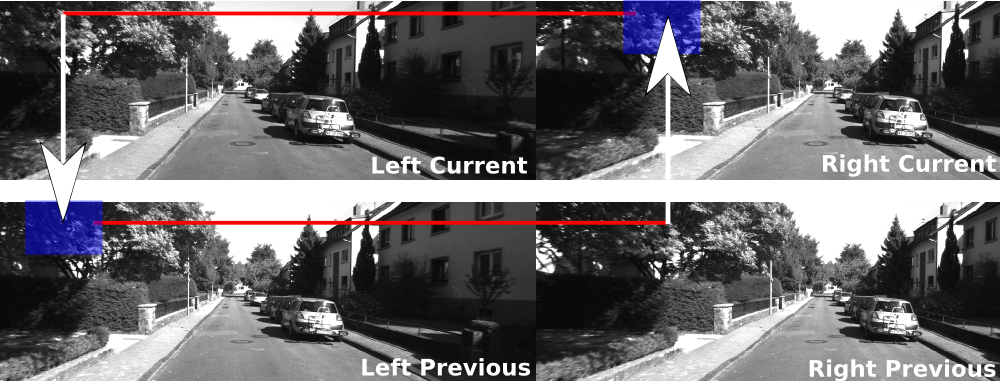}
\caption{The circular matching procedure which determines if a feature should be utilized for ego-motion estimation}
\label{fig:circular}
\end{figure}

%Ego-motion estimation
In order to perform ego-motion estimation, a variation of the method utilized by StereoScan \cite{c12} has been implemented. As in the StereoScan algorithm, features which are to be utilized in the ego-motion estimation step need to be `circularly' matched between the left and right current images at time $t$, as well as between the left and right images from the previous timestep $t-1$. Our procedure is performed as seen in Figure \ref{fig:circular}. First, FREAK descriptors are matched between the current left and right stereo images. A single descriptor match, $d_{left, current}$ (the descriptor in the left image) and $d_{right, current}$ (the corresponding descriptor match in the right image), in Figure \ref{fig:circular} is used as an example. The descriptor in the current left image $d_{left, current}$ is then matched to the best matching descriptor in the previous left image $d_{left,prev, best}$ within a pre-defined window indicated by the large square. Once $d_{left,prev, best}$ is found, the corresponding descriptor match in the previous right image $d_{right, prev, best}$ is available as this has been computed in the previous step. A match between the previous right image descriptor $d_{right, prev, best}$ and the best current right image descriptor $d_{right, current, best}$ is then computed within a pre-defined window. If $d_{right, current, best}$ matches $d_{right, current}$, then a circular match is found. All circular matches are then utilized to perform ego-motion estimation.\\

StereoScan utilizes the epipolar constraint to efficiently match descriptors between the left and right images \cite{c12}. Our variation does not require the epipolar constraint to match features between images as feature descriptors have already been matched between images using the FREAK algorithm and therefore are instantly available.\\ 

The circular matches are then utilized to minimize the sum of reprojection errors using Gauss-Newton Optimization as performed in \cite{c12}. RANSAC is then performed $50$ times to ensure that the rotation matrix $R$ and translation vector $t$ are accurately estimated.

\section{Experiments and Results}
This algorithm has been tested on the KITTI dataset \cite{c16} and Tsukuba Dataset \cite{c18,c19}. KITTI is arguably the best available benchmark dataset for Visual Odometry. The dataset is comprised of images taken from a stereo mount placed on a car driving around Karlsruhe, Germany. The CFORB algorithm achieved very competitive results being ranked among the top $25$ VO algorithms on this dataset \footnote{KITTI Visual Odometry Rankings:\\ http://www.cvlibs.net/datasets/kitti/eval\_odometry.php}. On the Tsukuba dataset, CFORB achieved a competitive, average translational accuracy of $3.70 \%$. CFORB has been tested on an $8$ core, $3.0 GHz$ desktop.

\section{KITTI Performance}
After running CFORB on the entire KITTI dataset, the algorithm achieved an average translation accuracy of $3.73 \%$ and a rotational error of $0.0107 deg/m$. The average run-time of the algorithm is $1 Hz$ without optimization\footnote{The current code run-time can be improved upon by at least an order of magnitude}. We also decided to analyze the performance of CFORB on a highly textured subset of the KITTI dataset. The length of this subset sequence is approximately $2.5 km$ long. This specific sequence produced significantly better results than the overall average performance obtained from running the algorithm on the entire dataset. The reason for this performance improvement is discussed in the section to follow.

\subsection{Performance as a Function of Length}
\label{sec:length}
\begin{figure}[!t]
\centering
\includegraphics[width=3.0in]{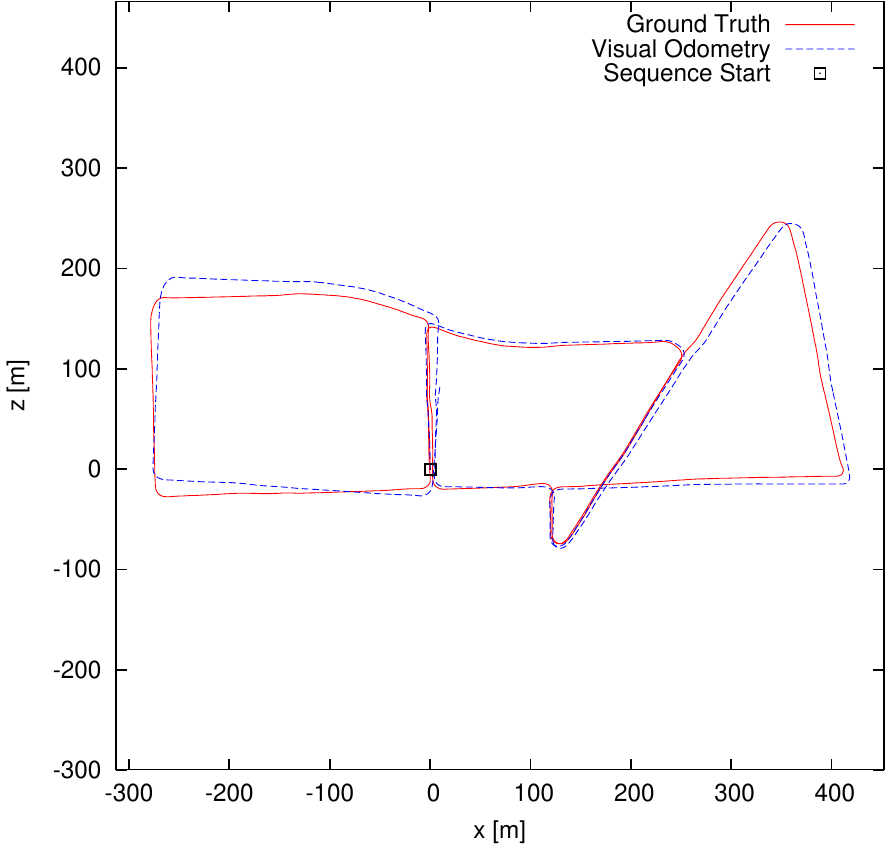}
\caption{CFORB run on a sequence of the KITTI dataset. The CFORB trajectory is indicated by the dashed lines}
\label{fig:00}
\end{figure}

\begin{figure}[!t]
\centering
\includegraphics[width=3.0in]{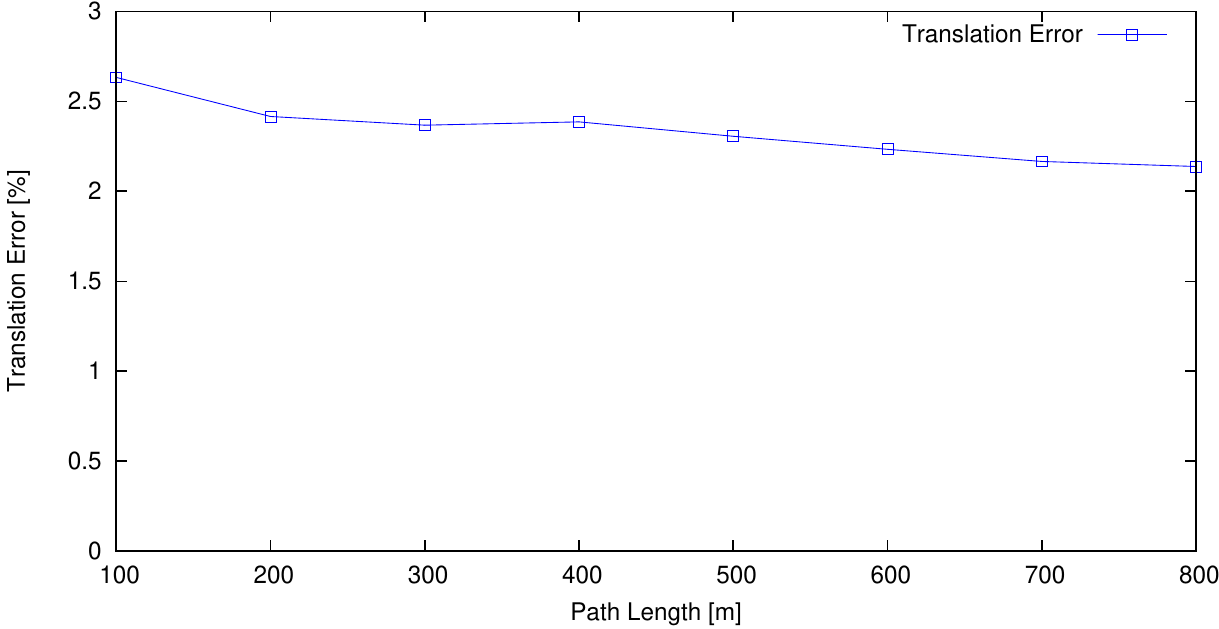}
\caption{The average translation error as a function of path length}
\label{fig:00tl}
\end{figure}

\begin{figure}[!t]
\centering
\includegraphics[width=3.0in]{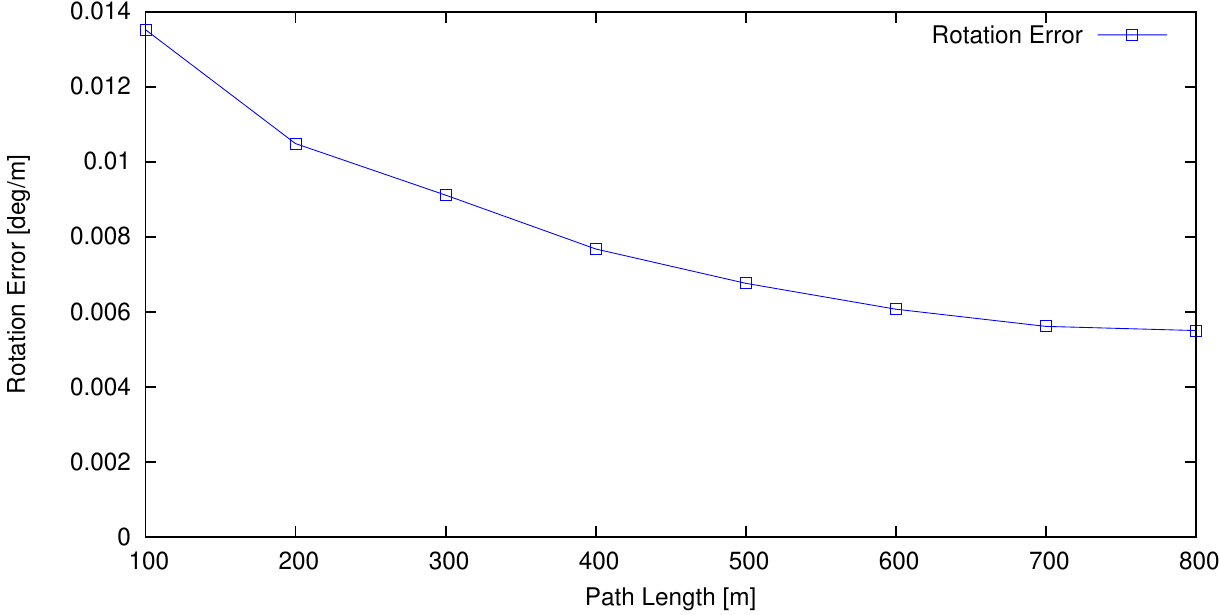}
\caption{The average rotation error as a function of path length}
\label{fig:00rl}
\end{figure}

We ran CFORB on the highly textured subset sequence from the KITTI dataset. CFORB generated the dashed line trajectory shown in Figure \ref{fig:00}. The average translation error as a function of length is $2.4 \%$ as has been calculated from Figure \ref{fig:00tl}. The average rotational error is  $0.009 deg/m$ and has been calculated from Figure \ref{fig:00rl}.\\

\begin{figure}[!t]
\centering
\includegraphics[width=2.5in]{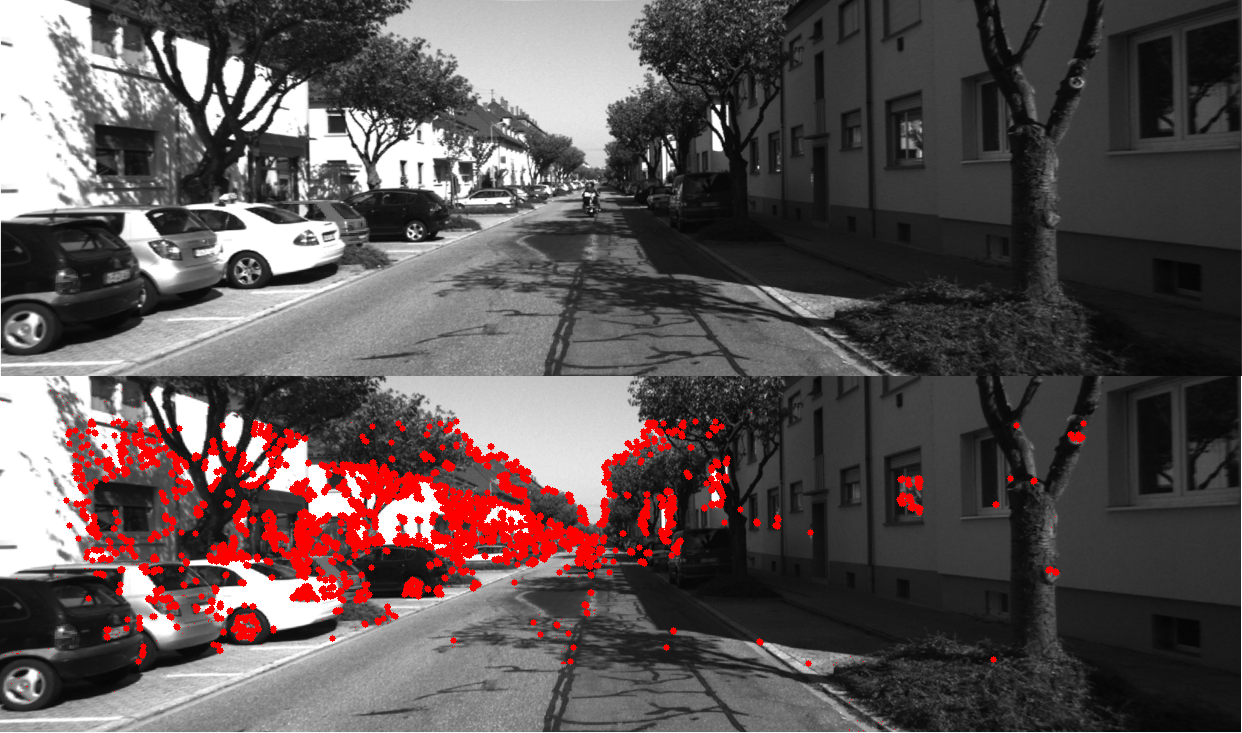}
\caption{The top image is the original image. The bottom image displays interest points, indicated by red dots, detected in a highly textured environment.}
\label{fig:textured}
\end{figure}

These results are significantly better than the average CFORB performance for a number of reasons. Firstly, if the dataset was generated in a highly textured environment, better performance is obtained as more unique and repeatable interest points are detected. This dataset is indeed highly textured and an example image with detected ORB features (indicated by red dots) can be seen in Figure \ref{fig:textured}. Many of the detected features corresponds to unique objects such as cars, the motorcycle in the center of the road and houses which are both unique and consistent in subsequent images. \\

In addition, if the detected features are uniformly spread around the image, then a better estimation of the ego-motion can be obtained \cite{c12}. It is therefore possible to detect features with a uniform spread across the image using techniques such as feature bucketing.

\subsection{Performance as a Function of Speed}
\label{sec:speed}
\begin{figure}[!t]
\centering
\includegraphics[width=3.0in]{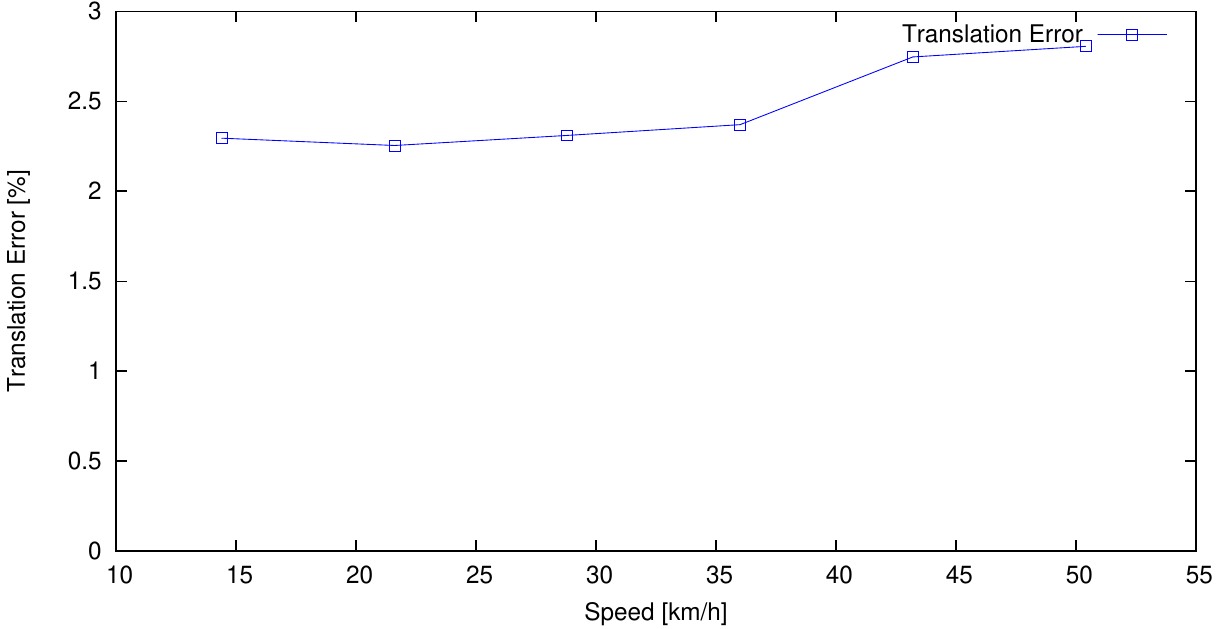}
\caption{The average translation error as a function of speed}
\label{fig:00ts}
\end{figure}

\begin{figure}[!t]
\centering
\includegraphics[width=3.0in]{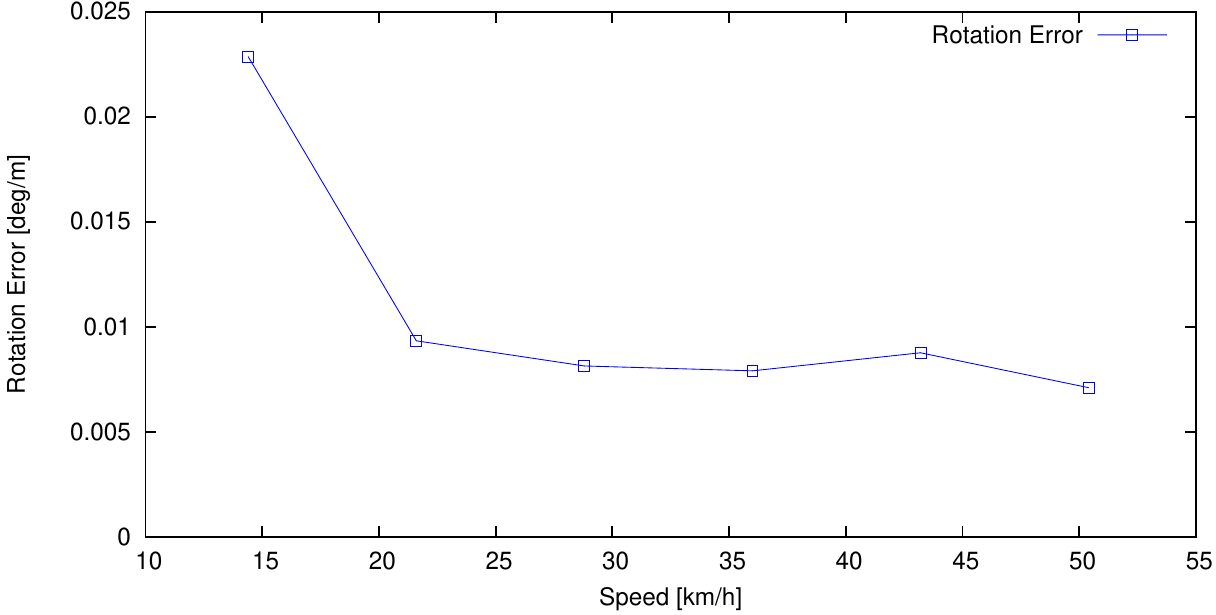}
\caption{The average rotation error as a function of speed}
\label{fig:00rs}
\end{figure}

The performance of CFORB on the above-mentioned KITTI sequence has also been tabulated as a function of the speed at which the stereo-rig is travelling. The average translation error as a function of speed is $2.5 \%$, calculated from Figure \ref{fig:00ts}, and the average rotational error is $0.013 deg/m$, calculated from Figure \ref{fig:00rs}. As seen in Figure \ref{fig:00ts}, CFORB is robust to changes in speed as the average translational error is consistently between $2-3 \%$ regardless of the speed at which the stereo-rig is travelling. The performance degrades slightly at higher speeds. This may be due to the fact that less repeatable features are detected at high speeds due to factors such as motion blur. The rotational error is consistent at speeds greater than $20 km/h$. At lower speeds, the car tends to turn corners more frequently resulting in larger changes in orientation. At higher speeds the car does not drastically change its orientation resulting in better rotation estimates.

\subsection{Tsukuba Performance}
\label{sec:tsukuba}
\begin{figure}[!t]
\centering
\includegraphics[width=3.0in]{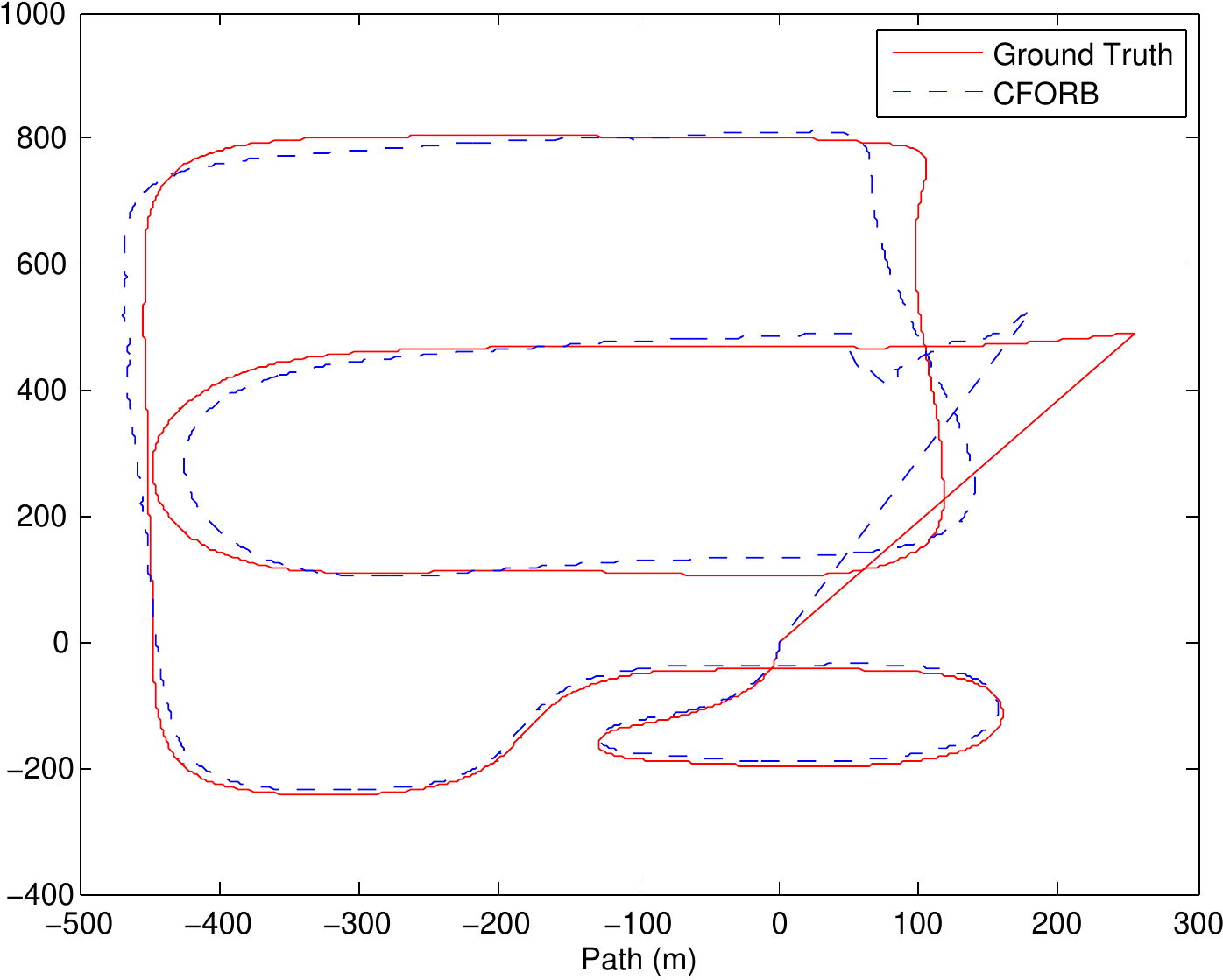}
\caption{The path trajectory compared to the ground truth for the Tsukuba dataset}
\label{fig:tsukubapath}
\end{figure}

As mentioned in Section \ref{sec:introduction}, indoor localization is crucial in many applications ranging from inventory management to robotic tour guides. We therefore decided to test the algorithm on an indoor dataset in order to verify the robustness of the algorithm to these types of environments. The algorithmic performance on the Tsukuba dataset achieved a competitive average translation accuracy of $3.70 \%$. This is comparable to outdoor datasets as the average translational accuracy on the entire KITTI dataset is $3.73 \%$. The path trajectory compared to the ground truth is shown in Figure \ref{fig:tsukubapath}. This is highly accurate and indicates that the algorithm is indeed suitable for use in indoor environments. In addition, this algorithm is robust to camera movements over uneven terrain. An example of this is seen in Figure \ref{fig:tsukubapath} where matches between a stereo image pair from the Tsukuba dataset are shown. The stereo-rig moves unevenly through the environment and still manages to detect and match features between the images.

%\begin{figure}[!t]
%\centering
%\includegraphics[width=2.0in]{../Drawings/comparison.pdf}
%\caption{A comparison between the outdoor and indoor performance of the CFORB algorithm}
%\label{fig:comparison}
%\end{figure}

\begin{figure}[!t]
\centering
\includegraphics[width=3.0in]{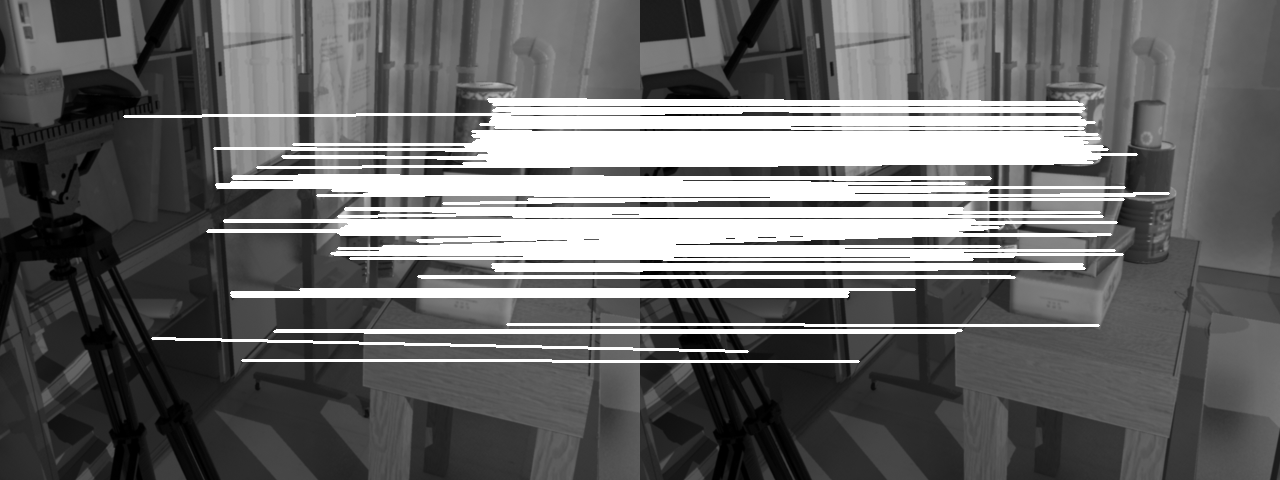}
\caption{An example of matching between images in the Tsukuba dataset where the cameras are moving over an uneven indoor terrain}
\label{fig:comparison}
\end{figure}

%\subsection{Malaga Performance}
%\label{sec:tsukuba}

%\begin{figure}[!t]
%\centering
%\includegraphics[width=3.5in]{../Drawings/malaga/malaga_tl.png}
%\caption{The translation error as a function of path length on the Tsukuba dataset}
%\label{fig:tsukubatl}
%\end{figure}

\section{Discussion}
We have presented the novel Circular FREAK-ORB (CFORB) Visual Odometry (VO) algorithm. This algorithm detects features using the ORB algorithm and computes FREAK descriptors. This algorithm is invariant to both rotation and scale. Invalid matches are removed using the vertical and horizontal geometric constraints which have not previously been utilized in a VO algorithm. CFORB also implements a variation of circular matching \cite{c12} which removes the need to match features between images using the epipolar constraint. This algorithm exhibits competitive performance on the KITTI dataset achieving an average translational accuracy of $3.73\%$ and an average rotational accuracy of $0.0107 deg/m$. It is among the top $25$ algorithms on the KITTI VO rankings. We tested the algorithm in a highly textured environment with an approximately uniform feature spread across the images and achieved an average translation error of $2.4 \%$ and an average rotational error of $0.009 deg/m$. In addition, this algorithm is suitable for indoor applications achieving an average translational error on the Tsukuba dataset of $3.70 \%$. The algorithm is also robust to uneven terrain as shown in Section \ref{sec:tsukuba}.

\addtolength{\textheight}{-12cm}   % This command serves to balance the column lengths

%%%%%%%%%%%%%%%%%%%%%%%%%%%%%%%%%%%%%%%%%%%%%%%%%%%%%%%%%%%%%%%%%%%%%%%%%%%%%%%%

%%%%%%%%%%%%%%%%%%%%%%%%%%%%%%%%%%%%%%%%%%%%%%%%%%%%%%%%%%%%%%%%%%%%%%%%%%%%%%%%

%%%%%%%%%%%%%%%%%%%%%%%%%%%%%%%%%%%%%%%%%%%%%%%%%%%%%%%%%%%%%%%%%%%%%%%%%%%%%%%%
%\section*{APPENDIX}
%\section*{ACKNOWLEDGMENT}
%
%The preferred spelling of the word ÒacknowledgmentÓ in America is without an ÒeÓ after the ÒgÓ. Avoid the stilted expression, ÒOne of us (R. B. G.) thanks . . .Ó  Instead, try ÒR. B. G. thanksÓ. Put sponsor acknowledgments in the unnumbered footnote on the first page.

%%%%%%%%%%%%%%%%%%%%%%%%%%%%%%%%%%%%%%%%%%%%%%%%%%%%%%%%%%%%%%%%%%%%%%%%%%%%%%%%

%References are important to the reader; therefore, each citation must be complete and correct. If at all possible, references should be commonly available publications.

\end{document}